\documentclass[runningheads]{llncs}
\usepackage{graphicx}
\usepackage{amsmath,amssymb} % define this before the line numbering.
\usepackage{color}
\usepackage{epigraph}
\usepackage{algorithmicx}
\usepackage{algorithm}
\usepackage{algpseudocode}

\DeclareMathOperator*{\argmax}{arg\!max}
\DeclareMathOperator{\Deg}{deg}
\DeclareMathOperator{\Dist}{dist}
\DeclareMathOperator{\Normalize}{normalize}

\begin{document}
\title{Visual Psychophysics for Making Face Recognition Algorithms More Explainable}

\titlerunning{Visual Psychophysics for Face Recognition}

\author{Brandon~RichardWebster\inst{1}\orcidID{0000-0003-4278-1282} \and So~Yon~Kwon \and
Christopher~Clarizio\inst{1} \and Samuel~E.~Anthony\inst{2,3} \and Walter~J.~Scheirer\inst{1}}
% First Name: Brandon, Last Name: RichardWebster
% there is no space in RichardWebster, and W is capitalized
%
% First Name: So Yon, Last Name: Kwon

\authorrunning{RichardWebster et al.}

\institute{University of Notre Dame, Notre Dame, IN, 46556, USA \and
Perceptive Automata, Inc. \and
Harvard University, Cambridge, MA 02138, USA}

\maketitle

\begin{abstract}
Scientific fields that are interested in faces have developed their own sets of concepts and procedures for understanding how a target model system (be it a person or algorithm) perceives a face under varying conditions. In computer vision, this has largely been in the form of dataset evaluation for recognition tasks where summary statistics are used to measure progress. While aggregate performance has continued to improve, understanding individual causes of failure has been difficult, as it is not always clear why a particular face fails to be recognized, or why an impostor is recognized by an algorithm. Importantly, other fields studying vision have addressed this via the use of visual psychophysics: the controlled manipulation of stimuli and careful study of the responses they evoke in a model system. In this paper, we suggest that visual psychophysics is a viable methodology for making face recognition algorithms more explainable. A comprehensive set of procedures is developed for assessing face recognition algorithm behavior, which is then deployed over state-of-the-art convolutional neural networks and more basic, yet still widely used, shallow and handcrafted feature-based approaches. {\renewcommand{\thefootnote}{\fnsymbol{footnote}} \footnotetext[1]{Funding was provided under IARPA contract \#D16PC00002, NSF DGE \#1313583, and NSF SBIR Award \#IIP-1738479. Hardware support was generously provided by the NVIDIA Corporation, and made available by the National Science Foundation (NSF) through grant \#CNS-1629914.}}   

\keywords{Face Recognition, Biometrics, Explainable AI, Visual Psychophysics, Biometric Menagerie}
\end{abstract}

\setcounter{footnote}{0} 

\section{Introduction}
\label{sec:intro}
With much fanfare, Apple unveiled its Face ID product for the iPhone X in the Fall of 2017 at what was supposed to be a highly scripted event for the media. Touted as one of the most sophisticated facial recognition capabilities available to consumers, Face ID was designed to tolerate the wide range of user behaviors and environmental conditions that can be expected in a mobile biometrics setting. Remarkably, during the on-stage demo, Face ID failed~\cite{apple}. Immediate speculation, especially from those with some familiarity with biometrics, centered around the possibility of a false negative, where an enrolled user failed to be recognized. After all, it was very dark on stage, with a harsh spotlight on the presenter, whose appearance was a bit more polished than usual --- all variables that conceivably were not in the training set that the deep learning-based model behind Face ID was trained on. Apple, for its part, released a statement claiming that it was too many imposter authentication attempts before the demo that caused the problem~\cite{hern_2017}. Of course, that did little to satisfy the skeptics. 

\begin{figure*}[t]
 \centering
    \includegraphics[width=\textwidth]{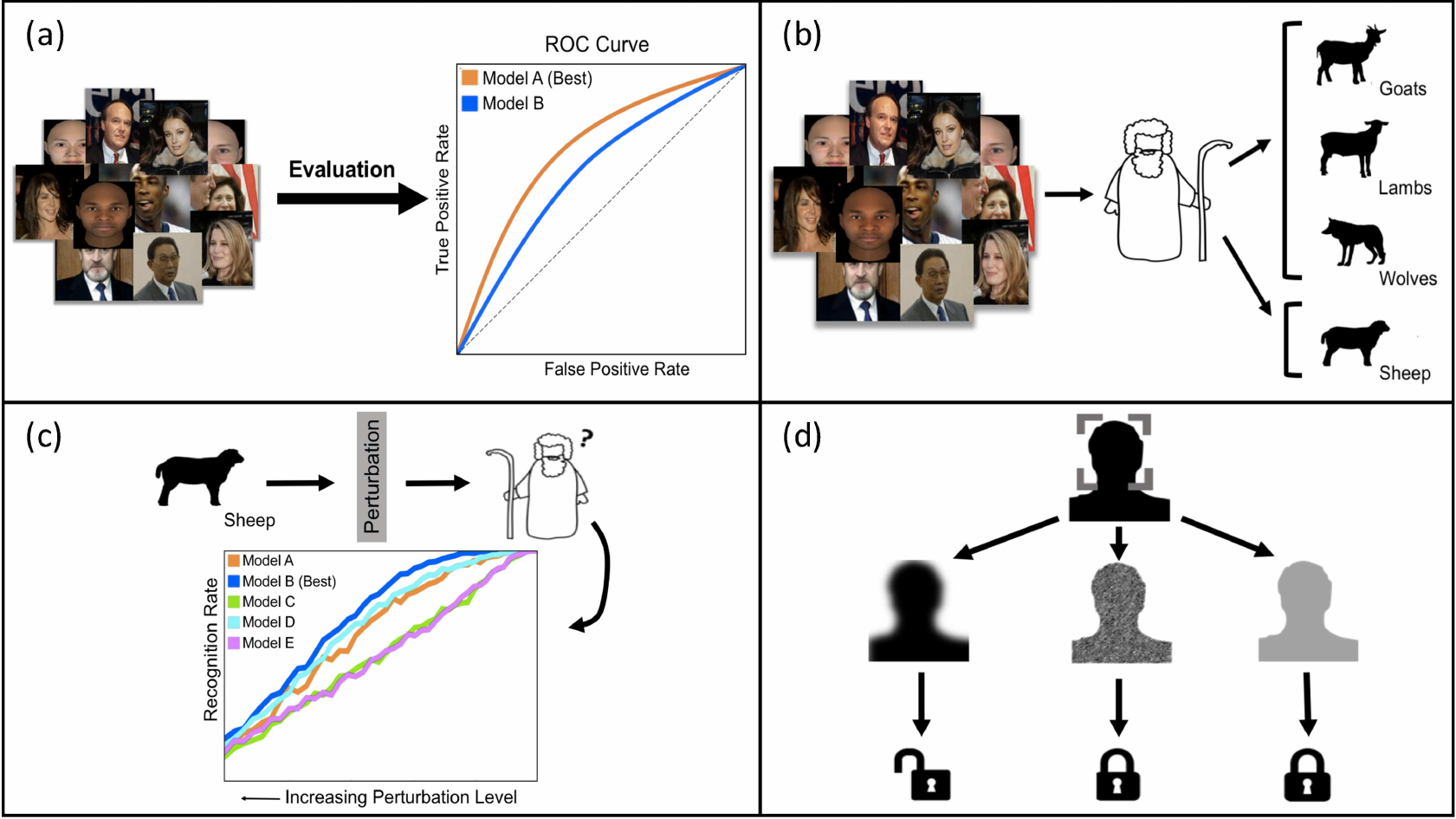}
 \caption{Visual Pyschophysics~\cite{embretson2000item,lu2013visual,prins2016psychophysics} helps us explain algorithm behavior in a way that traditional dataset evaluation (a) cannot. Our proposed methodology introduces a theoretical mapping between elements of psychopysical testing and the biometric menagerie paradigm~\cite{4711054}, where a shepherd function first isolates cooperative users (``sheep") from all others (b). From a perfect matching scenario, the images of the sheep are incrementally perturbed using a chosen image transformation, and item-response curves are plotted so that points of failure can be identified (c). The results can then be used to explain why matching works for some input images, but not others (d).}
 \label{fig:teaser}
\end{figure*}

This controversy highlights a critical difficulty now facing the computer vision community: what is the true source of a problem when the object of study is a black box? While Apple may have access to the internals of its phones, ordinary users do not. But even with direct access to an algorithm, we can't always get what we want when it comes to an understanding of the conditions that lead to failure~\cite{hendricks2016generating,ribeiro2016should}. Given the fact that face recognition is one of the most common user-facing applications in computer vision, the ability to diagnose problems and validate claims about algorithm design and performance is desirable from the perspective of both the researcher and administrator charged with operating such systems. This is exactly why we want AI for face recognition to be \textit{explainable}. In this paper, we look at a new methodology for doing this with any face recognition algorithm that takes an image as input. But first, let us consider the way we currently use evaluation procedures to try to understand the output of face recognition systems.
    
The development cycle of face recognition algorithms relies on large-scale datasets. Progress is measured in a dataset context via summary statistics (\textit{e.g.}, false positive rate, true positive rate, identification rate) computed over an evaluation set or $n$ folds partitioned~\cite{haralick1992performance} from the evaluation set and expressed as a ROC or CMC curve (Fig.~\ref{fig:teaser}, Panel a). Such datasets have become even more important with the rise of machine learning, where both large training and evaluation sets are needed. For face verification (1:1 matching), there are a number of datasets that brought performance up to usable levels in controlled settings with cooperative subjects~\cite{phillips2000feret,phillips2005overview,phillips2009overview,beveridge2013challenge,phillips2011introduction}. More recently, web-scale data~\cite{kemelmacher2016megaface,Klare_2015_CVPR,yi2014learning,ortiz2014face,bhattarai2014some,schroff2015facenet,wang2017face} has been used to investigate more difficult recognition settings including face identification (1:$N$ matching) and challenging impostor settings. There is a continuing push for larger datasets, which does not always address the problems observed in the algorithms trained over them. While aggregate performance has continued to improve, understanding individual causes of failure remains difficult, as it is not always clear why a particular face fails to be recognized, or why an impostor is recognized by an algorithm when considering a summary statistic.     
    
Importantly, other fields studying vision have addressed this via the use of visual psychophysics: the controlled manipulation of stimuli and careful study of the responses they evoke in a model system~\cite{embretson2000item,lu2013visual,prins2016psychophysics}. In particular, the field of psychology has developed specific concepts and procedures related to visual psychophysics for the study of the human face and how it is perceived~\cite{tanaka1993parts,duchaine2006cambridge,oosterhof2008functional,germine2011cognitive}. Instead of inferring performance from summary statistics expressed as curves like ROC or CMC, visual psychophysics allows us to view  performance over a comprehensive range of conditions, permitting an experimenter to pinpoint the exact condition that results in failure. The gold standard for face recognition experimentation with people is the Cambridge Face Memory Test~\cite{duchaine2006cambridge}, which uses progressively degraded variations of faces to impede recognition. It has led to landmark studies on prosopagnosia (the inability to recognize a face)~\cite{duchaine2007family}, super recognizers (people with an uncanny ability to recognize faces)~\cite{russell2009super}, and face recognition ability and heritability~\cite{wilmer2010human}. Similarly, visual psychophysics has been used to study the role of holistic features in recognition by swapping parts to break the recognition ability~\cite{tanaka1993parts}. More recent work has moved into the realm of photo-realistic 3D face synthesis, where changes in face perception can be studied by varying aspects of facial anatomy~\cite{oosterhof2008functional} and the age of the face used as a stimulus~\cite{germine2011cognitive}. Given the breadth of its applicability, psychophysics also turns out to be an extremely powerful regime for explaining the behavior of algorithms.  

We already see visual psychophysics becoming an alternate way of studying algorithm behavior in other areas of computer vision such as object recognition~\cite{richardwebster2016psyphy}, face detection~\cite{Scheirer_2014_TPAMIa}, and reinforcement learning~\cite{leibo2018psychlab}. However, no work has been undertaken yet in the area of face recognition.
In this paper, we propose to address this by building a bridge from vision science to biometrics. Working from a recently established framework for conducting psychophysics experiments on computer vision algorithms~\cite{richardwebster2016psyphy} and infusing it with the proper methods from visual psychophysics for the study of face recognition in people, we fill in the missing pieces for automatic face recognition. Specifically, this involves a theoretical mapping between elements of psychopysical testing and the biometric menagerie paradigm~\cite{4711054}, where cooperative users (``sheep") are isolated (Fig.~\ref{fig:teaser}, Panel b), and incremental perturbations degrade their performance (Fig.~\ref{fig:teaser}, Panel c). Results gathered from psychophysics experiments making use of highly controlled procedurally generated stimuli can then inform the way we should use a face recognition algorithm by explaining its failure modes (Fig.~\ref{fig:teaser}, Panel d). 

\section{Related Work}
\label{sec:related}
\textbf{Explainable AI.} An increasing emphasis on artificial neural networks in AI has resulted in a corresponding uptick in interest in explaining how trained models work. With respect to representations, Zeiler and Fergus~\cite{zeiler2014visualizing} suggested that a multi-layer deconvolutional network can be used to project feature activations of a target convolutional network (CNN) back to pixel-space, thus allowing a researcher to reverse engineer the stimuli that excite the feature-maps at any layer in the CNN. Subsequent work by Mahendran and Vedaldi~\cite{Mahendran_2015_CVPR} generalized the understanding of representations via the analysis of the representation itself coupled with a natural image prior. With respect to decision making, Ribeiro et al.~\cite{ribeiro2016should} have introduced a framework for approximating any classifier with an explicitly  interpretable model. In a different, but related tactic, Fong et al.~\cite{Fong_2017_ICCV} use image perturbations to localize image regions relevant to classification. Image perturbations will form an important part of our methodology, described below in Sec.~\ref{sec:methods}. A number of alternative regimes have also been proposed, including a sampling-based strategy that can be applied to face recognition algorithms~\cite{7738872}, sampling coupled with reinforcement learning~\cite{hendricks2016generating}, and a comprehensive probabilistic programming framework~\cite{lake2015human}. What we propose in this paper is not meant to be a replacement for any existing method for explaining an AI model, and can work in concert with any of the above methods.
    
\textbf{Psychophysics for Computer Vision.} The application of psychophysics to computer vision has largely been an outgrowth of interdisciplinary work between brain scientists and computer scientists looking to build explanatory models that are consistent with observed behavior in animals and people. A recent example of this is the work of Rajalingham et al.~\cite{rajalingham2018large}, which compares the recognition behavior of monkeys, people and CNNs, noting that CNNs do not account for the image-level behavioral patterns of primates. Other have carried out studies using just humans as a reference point, with similar conclusions~\cite{gerhard2013sensitive,eberhardt2016deep,DBLP:journals/corr/GeirhosJSRBW17,heath1996comparison}. With respect to approaches designed specifically to perform psychophysics on computer vision algorithms, a flexible framework is PsyPhy, introduced by RichardWebster et al.~\cite{richardwebster2016psyphy}. PysPhy facilitates a psychophysical analysis for object recognition through the use of item-response theory. We build from that work to support a related item-response analysis for face recognition. Outside of research to explain the mechanisms of AI algorithms, other work in computer vision has sought to infuse psychophysical measurements into machine learning models~\cite{Scheirer_2014_TPAMIa,McCurie_2017_FG}. Data in several of these studies has relied on the popular crowdsourced psychophysics website TestMyBrain.org~\cite{germine2012web}. In this work, we make use of a similar human-testing platform for comparison experiments.

\textbf{Methods from Psychology Applied to Biometrics.} While there is growing interest in what psychology can teach computer vision at large, the biometrics community was early to adopt some of its methods. Sinha et al.~\cite{sinha2006face} outlined 19 findings from human vision that have important consequences for automatic face recognition. Several of these findings have served as direct inspiration for the adoption of CNNs for face recognition. A significant outgrowth of NIST-run face recognition evaluations has been a series of human vs. computer performance tests~\cite{o2007face,o2008humans,o2012comparing,phillips2014comparison,phillips2015human}. Even though these studies have not made use of psychophysics, they still shed new light on face recognition capabilities. In some cases such as changes in illumination~\cite{o2007face,o2008humans}, good quality images~\cite{o2012comparing}, and matching frontal faces in still images~\cite{phillips2014comparison}, algorithms have been shown to be superior. However, one should keep in mind that these are controlled (or mostly controlled) verification settings, where images were intentionally acquired to reflect operational matching scenarios. In other cases, especially with more naturalistic data and video matching scenarios~\cite{phillips2014comparison,phillips2015human}, humans are shown to be superior. Studies such as these have established human perception as a measureable baseline for evaluating face recognition algorithms. We also look at human vs. algorithm performance as a baseline in this paper.

\textbf{Biometrics and Perturbed Inputs.} Many studies have sought to simulate real-world conditions that reduce matching performance. This has often taken the form of perturbations applied to the pixels on a face image --- the primary form of transformation we will consider for our psychophysics experiments. Karahan et al.~\cite{karahan2016image} and Grm et al.~\cite{8244393} have studied the impact of incrementally perturbing face images for transformations like Gaussian blur, noise, occlusion, contrast and color balance. In order to compensate for Gaussian blur, Ding and Tao~\cite{ding2017trunk} perturb sequences of face images for the purpose of learning blur-insensitive features within a CNN model. These experimental studies share an underlying motivation with this work, but are qualitatively and quantitatively different from the item-response-based approach we describe. 

\section{Psychophysics for Face Recognition Algorithms}
\label{sec:methods}

In the \textit{$M$-alternative forced-choice match-to-sample} ($M$-AFC) psychophysics procedure in psychology~\cite{prins2016psychophysics}, a \textit{sample} stimulus (\textit{e.g.}, visual, auditory, or tactile) is used to elicit a perceptual response from a subject. The subject is then given a refractory period to allow their response to return to neutral. Once their response returns to neutral, the subject is presented with an \textit{alternate} stimulus and given, if needed, another refractory period. This process is then repeated for a total of $M$ unique alternate stimuli. Finally, the subject is \textit{forced} to choose one of the alternate stimuli that best \textit{matched} the sample stimulus. This is where the procedure name $M$-alternative forced-choice match-to-sample comes from. By carefully linking sample or alternate stimuli to a single condition at a specific stimulus level, a scientist running the experiment can measure mean or median accuracy achieved at each of the observed stimulus levels across all subjects. Together, these stimulus levels and their aggregated accuracy yield an interpretable item-response curve~\cite{embretson2000item} (see Fig. 1, Panel c for an example).

RichardWebster et al.~\cite{richardwebster2016psyphy} introduced a technique using the $M$-AFC method to produce item-response curves for general object classification models that involves procedurally rendering objects. The process consists of two steps: (1) the identification of a preferred view and (2) the generation of an item-response curve. A preferred view is an extension of a canonical view~\cite{blanz1999object}, the theory that humans naturally prefer similar inter-class object orientations when asked for the best orientation which maximizes discriminability. The preferred view serves as the initial orientation of the procedurally rendered objects, allowing transformations such as rotation or scaling to guarantee a degradation of model performance. When item-response curves are generated, a modified $M$-AFC procedure is invoked that maps the alternate choices to the output of a classifier. However, instead of explicitly presenting alternate choices, the alternate choices are implicitely the learned classes of the classifier. Thus accuracy is computed by how frequently the correct class was chosen.

Although psychophysics for face recognition uses the same foundational $M$-AFC match-to-sample concepts, in practice it is very different than the psychophysics procedure for  general object recognition. To begin with, an individual trial of the $M$-AFC procedure described above for human subjects is identical to the face identification procedure of biometrics. A face is acquired, and the system is queried to determine the identity of the face by matching the acquired image to enrolled faces within the system. Thus, a single $M$-AFC match-to-sample trial is equivalent to 1:$N$ identification in biometrics. However, one difference between an algorithm performing 1:$N$ matching and a human performing the same task is the need to set a threshold for the decision of ``match" or ``non-match" in the case of the algorithm (to reject match instances with insufficiently high scores). 

Like any good scientific method, a method from psychophysics  attempts to isolate a single variable to observe the effect it has on the rest of the system. In psychophysics experiments for face recognition, we call the isolated variable the perturbation level, which represents the degree of transformation applied with a perturbation function directly to an identity or to the image containing an identity. Thus, the first step in performing psychophysics for face recognition systems is to remove identities from an initial dataset that consistently cause false matches or false non-matches --- errors that are already inherent within the matching process and would be a confound to studying the effect of the transformation. Doddington et al.~\cite{doddington1998sheep} formally grouped users interacting with a biometric system into four classes whimsically named after farm animals, which together are called the \textit{biometric menagerie}~\cite{teli2011biometric,4711054}. The biometric menagerie consists of \textit{goats} (identities that are difficult to match), \textit{lambs} (identities that are easily impersonated), \textit{wolves} (identities that impersonate easily), and finally \textit{sheep} (identities that match well to themselves but poorly to others). Since we want to remove all identities that lead to errors, we must remove the wolves, goats, and lambs. We call this the ``herding" process.

\begin{algorithm}[t]
    \begin{algorithmic}[1]
        \Require{$\Upsilon$, a ``shepherd" function for a face recognition algorithm}
        \Require{$I$, a set of input identities from a dataset}
        \State{$S \leftarrow \Upsilon(I, I)$} \Comment{similarity matrix}
        \State{$S \leftarrow \frac{(S + S^\intercal)}{2}$} \Comment{enforce symmetry}
        \State{$t_{h} \leftarrow$ optimize loss function $\lambda$ with TPE} \Comment{Hyperopt} \cite{bergstra-2011-tpe,bergstra-2013-hyperopt,bergstra-2015-hyperopt}
        \State{$I_{h} \leftarrow \lambda(S,t_{h})$} \Comment{the ``sheep" identities produced by $\lambda$}
        \Ensure{$t_{h}$, the optimal threshold to produce $I_{h}$}
        \Ensure{$I_{h}$, the ``sheep" identities isolated by the optimal threshold $t_{h}$}
    \end{algorithmic}
    \caption{$H(\Upsilon, I)$: a ``herding" function to isolate Doddington et al.'s~\cite{doddington1998sheep} sheep from the goats, lambs, and wolves}
        \label{alg:herder}
\end{algorithm}

The herding function, $H$ (Alg. \ref{alg:herder}), takes a set of input identities from an initial dataset, $I$, and a ``shepherd" function, $\Upsilon$, as input, and determines which identities $\Upsilon$ considers sheep. The $\Upsilon$ function is a wrapper function to a face recognition algorithm, $f$, and accepts two sets of identities: $I_{p}$ the probe set and $I_{g}$ the gallery set. It returns a standard similarity matrix where $I_{p}$ is row-wise and $I_{g}$ is column-wise. An example shepherd function can be seen in Alg. \ref{alg:dec}. During the herding step, the input set $I$ is split into $I_{p}$ and $I_{g}$, which are used as input to $\Upsilon$. The herding function itself is quite simple: it obtains a similarity matrix from the shepherd function, forces matrix symmetry, and then optimizes the loss function, $\lambda$ (Alg. \ref{alg:loss}), for $250$ iterations with Hyperopt's implementation of the Tree-structured Parzen Estimator (TPE) hyperparameter optimizer~\cite{bergstra-2011-tpe,bergstra-2013-hyperopt,bergstra-2015-hyperopt}. More complicated is the loss function $\lambda$ that the herding function uses.

$\lambda$ takes as input a similarity matrix, $S$, and a threshold, $t$. The first step, thresholding the matrix, is standard in biometrics applications. However, the next step is not. The thresholded matrix is then XORed with an identity matrix, $\mathcal{I}$, to isolate all of the false match and false non-match pairs of identities ($\mathcal{I}$ represents the correct true matches). This new matrix can be considered an adjacency matrix, $G$, where all of the edges represent the false matches and false non-matches and each vertex is an identity. 

The next step is to selectively remove vertices / identities until no edges remain while also removing as small a number of identities as possible. A strategy inspired by graph cuts  allows us to sort the vertices by degree, remove the vertex with the highest degree from $G$, and repeat until no edges in $G$ remain (see Supp. Alg. 1 for the exact description\footnote{Supp. mat. available at \url{http://www.bjrichardwebster.com/papers/menagerie/supp}}). At the end, $G$ will be a completely disconnected graph, where no remaining identity will cause a false match or false non-match with any other remaining identity. By definition, all of the remaining identities are sheep.
The returned loss value is the number of identities removed, where the function favors a lower false match rate, \textit{i.e.}, higher thresholds are favored. After $\lambda$ is optimized, the  optimal threshold $t_{h}$ and sheep identities $I_{h}$ are returned.

\begin{algorithm}[t]
    \begin{algorithmic}[1]
        \Require{$f$, a face recognition function that produces a feature representation}
        \Require{$I_{p}$, a set of probe identities}
        \Require{$I_{g}$, a set of gallery identities}
        \State{$R_{p} \leftarrow i\in I_{p}: f(i)$} \Comment{feature representation for each identity}
        \State{$R_{g} \leftarrow i\in I_{g}: f(i)$}
        \State{$S \leftarrow r_{p}\in R_{p}, r_{g}\in R_{g}: \Dist(r_{p},r_{g})$} \Comment{matrix of distances}
        \State{$S \leftarrow \Normalize(S)$} \Comment{normalize distances to standard similarity matrix}
        \Ensure{$S$, the similarity matrix}
    \end{algorithmic}
    \caption{$\Upsilon_{f}(I_{p}, I_{g})$: a ``shepherd" function that produces a similarity matrix for the face recognition function $f$}
        \label{alg:dec}
\end{algorithm}

\begin{algorithm}[t]
    \begin{algorithmic}[1]
        \Require{$S$, similarity matrix}
        \Require{$t$, a threshold}
        \State{$M \leftarrow S \geq t$}
        \State{$M \leftarrow M \oplus \mathcal{I}$} \Comment{isolate FM and FNM pairs}
        \State{$G = (V, E)$ from $M$} \Comment{adjacency list}
        \State{$\nu \leftarrow |V|$}
        \While{$|E| > 0$} \Comment{remove goats, lambs, and wolves}
            \State{$v_{r} \leftarrow \argmax_{v \in V} \Deg(v)$}
            \State{\textbf{remove} $v_{r}$ from $V$} \Comment{remove the vertex and connected edges from $G$}
        \EndWhile
        \State{$l \leftarrow \nu - |V|$} \Comment{number of goats, lambs, and wolves removed}
        \State{$l \leftarrow l + (1-0.99999*t)$} \Comment{favor lower FMR over FNMR}
        \Ensure{$l$, the loss value}
    \end{algorithmic}
    \caption{$\lambda(S, t)$: a loss function that favors more \textit{sheep}, and favors a lower false match rate (FMR) over false non-match rate (FNMR)}
        \label{alg:loss}
\end{algorithm}

The sheep identities $I_{h}$ and the threshold $t_{h}$ serve as two of the inputs to the item-response point generator function $\Phi$ (Alg. \ref{alg:point}). $\Phi$ generates a point on an item-response curve that represents the rank one match rate for a specific perturbation function, $T$, and its respective perturbation level.  The perturbation function takes an image and a perturbation level as input, applies some transformation to the image, and returns the transformed image. In the context of the biometric menagerie, this function is analogous to ``perturbing" a sheep (dying the wool, shearing the wool, etc.) and asking its shepherd if it can properly identify the sheep. Thus $\Phi$ also takes $\Upsilon$ as a parameter. $\Phi$ uses $T$ to perturb each input identity in $I_{h}$ to create the set of perturbed probe identities for 1:$N$ identification. The remaining steps of $\Phi$ are standard to face recognition  systems operating in the identification mode: obtain similarity matrix from probe to gallery pairs, threshold the matrix, and calculate the match rate. The return value of the $\Phi$ function is an $x,y$ coordinate pair $\{s,\alpha\}$ for one item-response point, where $s$ represents the perturbation level and $\alpha$ is the match rate.

\begin{algorithm}[t]
    \begin{algorithmic}[1]
        \Require{$\Upsilon$, a ``shepherd" function for a facial recognition model}
        \Require{$I_{h}$, the ``sheep" identities for the found threshold $t_{h}$}
        \Require{$t_{h}$, the optimal threshold to produce $I_{h}$}
        \Require{$\delta$, the stimulus level}
        \State{$I_{h}' \leftarrow i \in I_{h}: T(i, \delta)$} \Comment{perturb identities to create probes}
        \State{$S \leftarrow \Upsilon(I_{h}', I_{h})$} \Comment{similarity matrix}
        \State{$M \leftarrow S \geq t_{h}$}
        \State{$\alpha \leftarrow \frac{|M \wedge \mathcal{I}|}{|I_{h}|}$} \Comment{obtain match rate using identity matrix $\mathcal{I}$}
        \Ensure{$\{s,\alpha\}$, an $x,y$ coordinate pair (stimulus level, match rate)}
    \end{algorithmic}
    \caption{$\Phi_T(\Upsilon,I_{h}, t_{h},\delta)$: an item-response point generation function for any image transformation function $T(i,\delta)$}
        \label{alg:point}
\end{algorithm}

\begin{algorithm}[t]
    \begin{algorithmic}[1]
        \Require{$\Upsilon$, a ``shepherd" function for a facial recognition model}
        \Require{$t_{h}$, the optimal threshold to produce $I_{h}$}
        \Require{$I_{h}$, the ``sheep" identities for the found threshold $t_{h}$}
        \Require{$n$, the number of stimulus levels}
        \Require{$b_{l}$ and $b_{u}$, the lower and upper bound values of the stimulus levels}
        \State{\textbf{Let} $\Delta$ be $n$ log-spaced stimulus levels from $b_{l}$ to $b_{u}$}
        \State{$k \leftarrow \bigcup\limits_{\delta \in \Delta} \{\Phi_{T} (\Upsilon,I_{h},t_{h},\delta)\}$}
        \Ensure{$k$, the item-response curve}
    \end{algorithmic}
    \caption{${\cal C}_{T}(\Upsilon,I_{h},t_{h},n,b_{l},b{u})$: an item-response curve generation function for any type of ``shepherd" function}
        \label{alg:gen}
\end{algorithm}

A shepherd's behavior for a set of sheep identities can be represented with an item-response curve (a collection of points obtained from $\Phi$), which is an interpretable representation of the shepherd's behavior in response to perturbation. For biometric identification, the x-axis is a series of values that represent a perturbation level from the original sheep identities and the y-axis is the match rate. To produce the item-response curves, the function ${\cal C}$ (Alg. \ref{alg:gen}) is called once for each transformation type. ${\cal C}$ repeatedly calls a point generated with $\Phi$ (Alg. \ref{alg:point}) to create one point for each stimulus level from the least amount of perturbation, $b_{l}$, to the most, $b_{u}$ ($b_{l}$ are the non-transformed sheep identities). The parameter $n$ is the number of stimulus levels to be used to produce the points on the match-response curve and are typically log-spaced to give finer precision near the non-transformed sheep identities. The final parameter $w$ is the number of identities examined at each stimulus level where $w \in [1,|I_{h}|]$.

\section{Experiments}

Experiments were designed with four distinct objectives in  mind: (1) to survey the performance of deep CNNs and other alternative models from the literature; (2) to look more closely at a surprising finding in order to explain the observed model behavior; (3) to study networks with stochastic outputs, which are prevalent in Bayesian analysis; and (4) to compare human vs. algorithm performance. For all experiments, we made use of the following face recognition algorithms: VGG-Face~\cite{parkhi2015deep}, FaceNet~\cite{facenet}, OpenFace~\cite{amos2016openface}, a simple three-layer CNN trained via high-throughput search of random weights~\cite{cox2011beyond} (labeled ``slmsimple" below), and OpenBR 1.1.0~\cite{6712754}, which makes use of handcrafted features. For each of the networks, the final feature layer was used with normalized cosine similarity as the similarity metric\footnote{Source code is available at \url{www.bjrichardwebster.com/papers/menagerie/code}}. All used models were used as-is from their corresponding authors, with no additional fine-tuning. A complete set of plots for all experiments can be found in the supplemental material. 

\textbf{Data Generation}. The following transformations were  applied to 2D images from the LFW dataset~\cite{huang2007labeled}: Gaussian blur, linear occlusion, salt \& pepper  noise, Gaussian noise, brown noise, pink noise, brightness, contrast, and sharpness. Note that we intentionally chose LFW because state-of-the-art algorithms have reached ceiling performance on it. The psychophysics testing regime makes it far more difficult for the algorithms, depending on the chosen transformation. Each face recognition algorithm was asked to ``herd" $1000$ initial images before item-response curve generation. All algorithms except OpenBR recognized all the initial images as sheep (see Supp. Sect. 2 for a breakdown). For each transformation, we generated $200$ different log-spaced stimulus levels, using each algorithm's choice of sheep, to create a corresponding item-response curve. In all, this resulted in ${\sim}5.5$ \textit{million} unique images and ${\sim}13.7$ \textit{billion} image comparisons.

Inspired by earlier work in psychology \cite{oosterhof2008functional,germine2011cognitive,yildirim2015efficient} making use of the FaceGen software package~\cite{singular}, we used it to apply transformations related to emotion and expression. A complete list can be found in the supplemental material. Each face algorithm selected sheep from $220$ initial images (all face textures provided by FaceGen, mapped to its ``average" 3D ``zero" model) for item-response curve generation. All chose $206$ sheep, with a nearly identical selection by each (see Supp. Sect. 3 for a complete list). $50$ stimulus levels were rendered for each image, resulting in ${\sim}400,000$ unique 3D images and ${\sim}17.5$ \textit{billion} image comparisons.

\begin{figure*}[t]
 \centering
    \includegraphics[width=\textwidth]{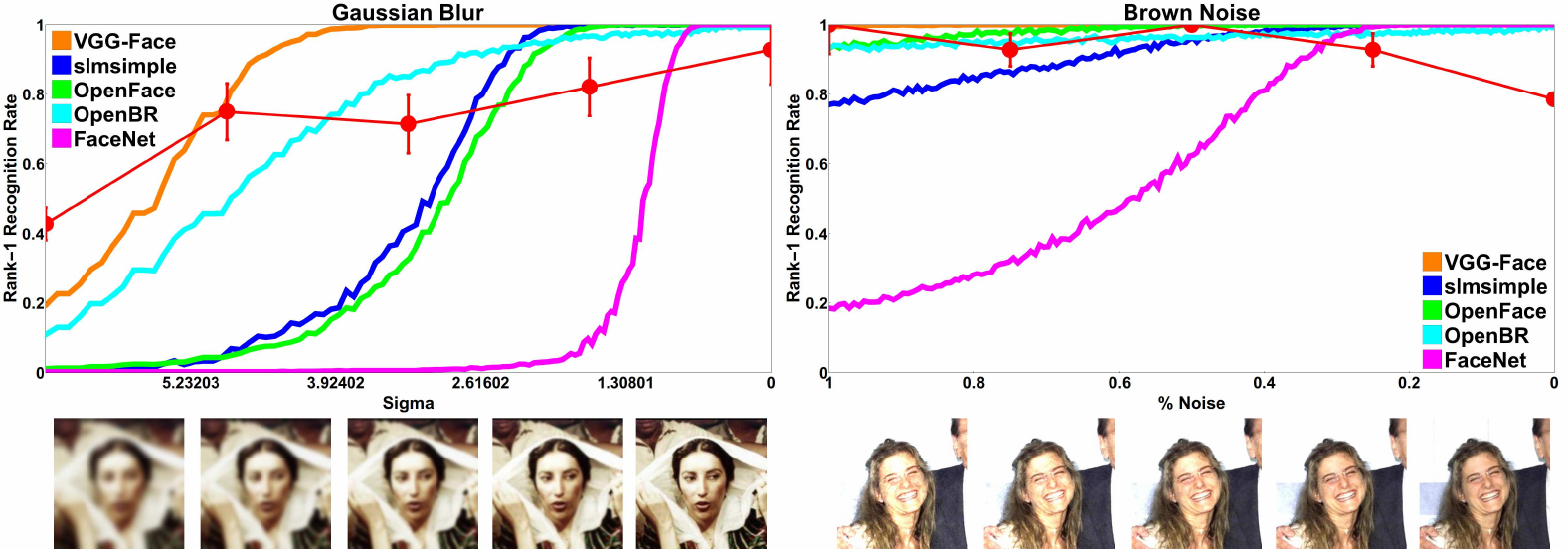}
 \caption{A selection of item-response curves for the $M$-AFC task using data from the LFW dataset~\cite{huang2007labeled}. Each experiment used five different face recognition algorithms~\cite{parkhi2015deep,facenet,amos2016openface,cox2011beyond,6712754}. A perfect curve would be a flat line at the top of the plot. The images at the bottom of each curve show how the perturbations increase from right to left, starting with no perturbation (\textit{i.e.}, the original image) for all conditions. The red dots indicate mean human performance for a selected stimulus level; error bars are standard error. Curves are normalized so chance is 0 on the y-axis. All plots are best viewed in color.}
 \label{fig:lfw}
\end{figure*}

\begin{figure*}[t]
 \centering
    \includegraphics[width=\textwidth]{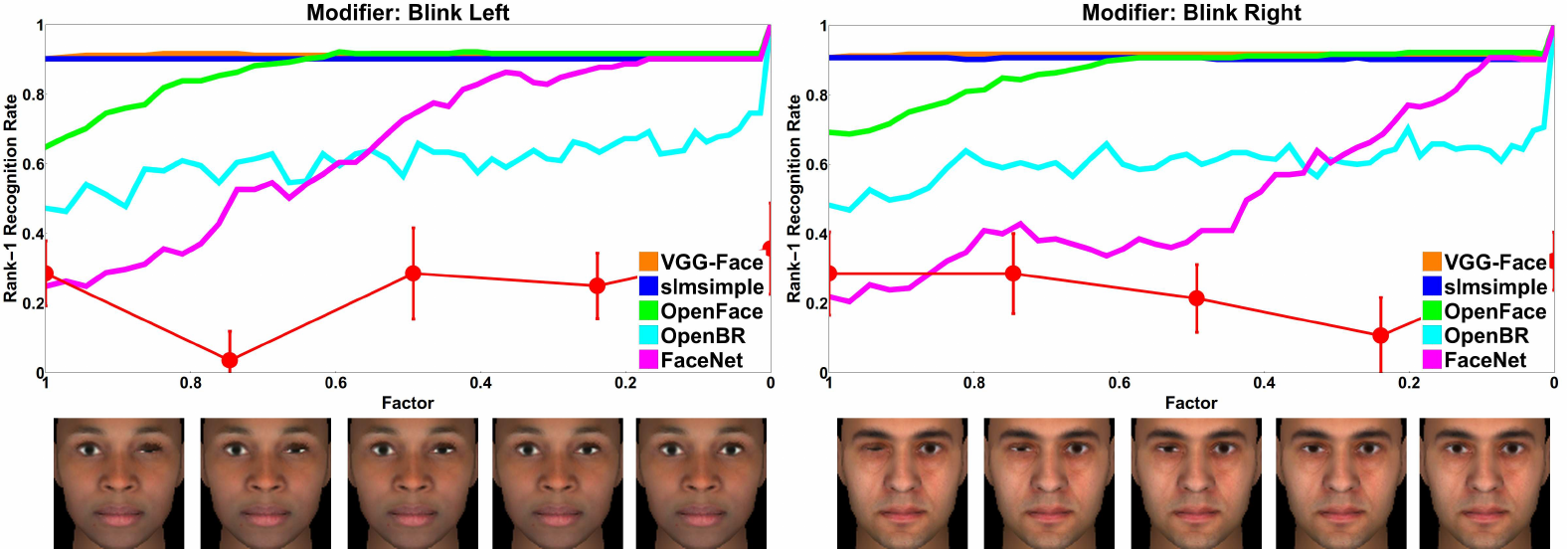}
 \caption{A selection of item-response curves for the $M$-AFC task using rendered 3D face models as stimuli~\cite{singular}. Curves are normalized so chance is 0. Here we see that three of the algorithms are drastically affected by the simple bodily function of blinking, while two others are not impacted at all. As in Fig.~\ref{fig:lfw}, VGG-Face is once again the best performing algorithm, but remarkably, we see that the three-layer CNN trained via a random search for weights (labeled ``slmsimple") works just as well.}
 \label{fig:fg}
\end{figure*}

\begin{figure*}[t]
 \centering
    \includegraphics[width=\textwidth]{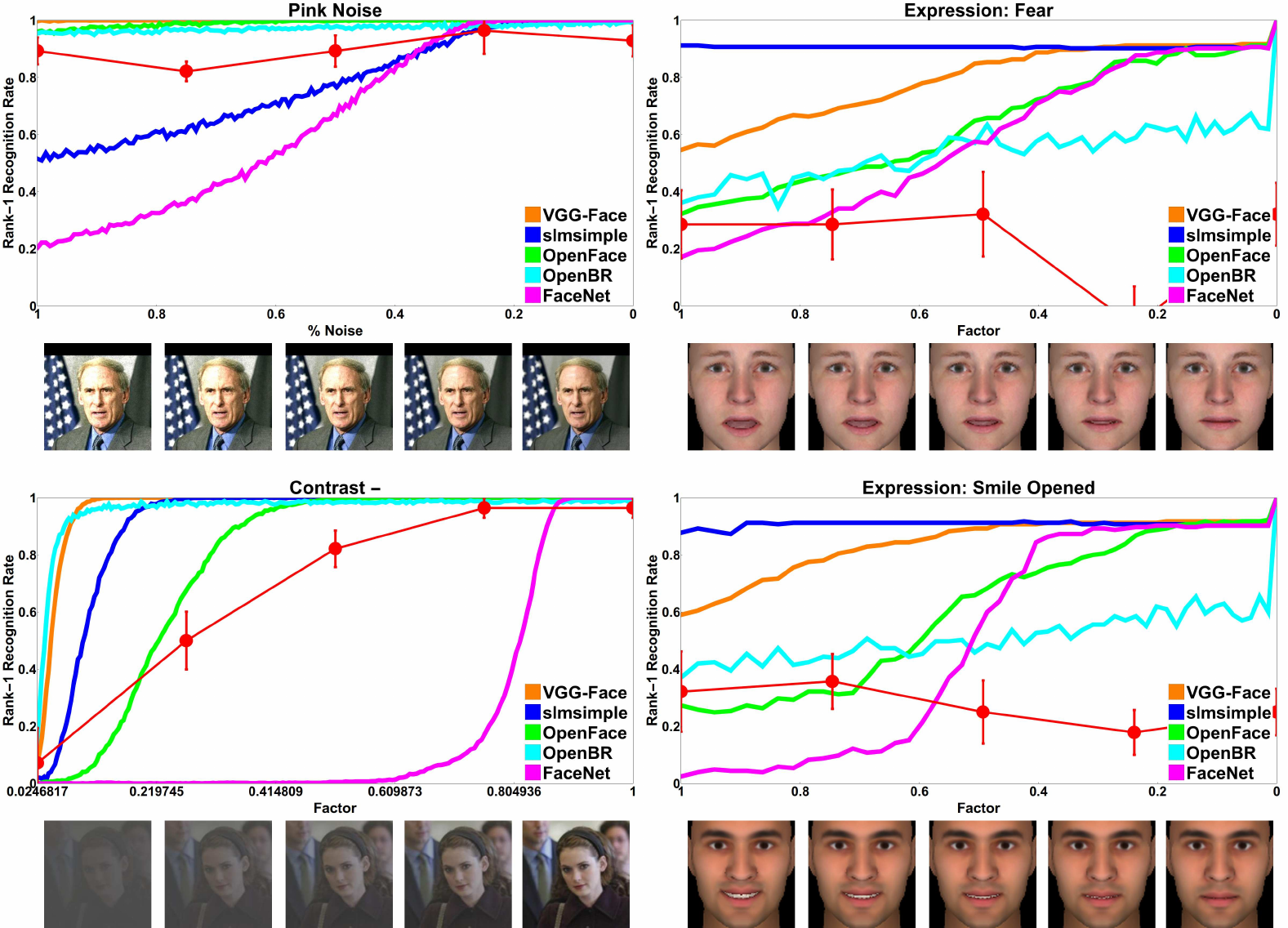}
 \caption{Two of the algorithms we evaluated, FaceNet~\cite{facenet} and OpenFace~\cite{amos2016openface}, both reported to be an implementation of Google's FaceNet~\cite{schroff2015facenet} algorithm. Curiously, we found major disagreement between them in almost all of our experiments. Note the gaps between their respective curves in the above plot. This performance gap was not evident when analyzing their reported accuracy performance on LFW.}
 \label{fig:composite}
\end{figure*}

\begin{figure*}[t]
 \centering
    \includegraphics[width=\textwidth]{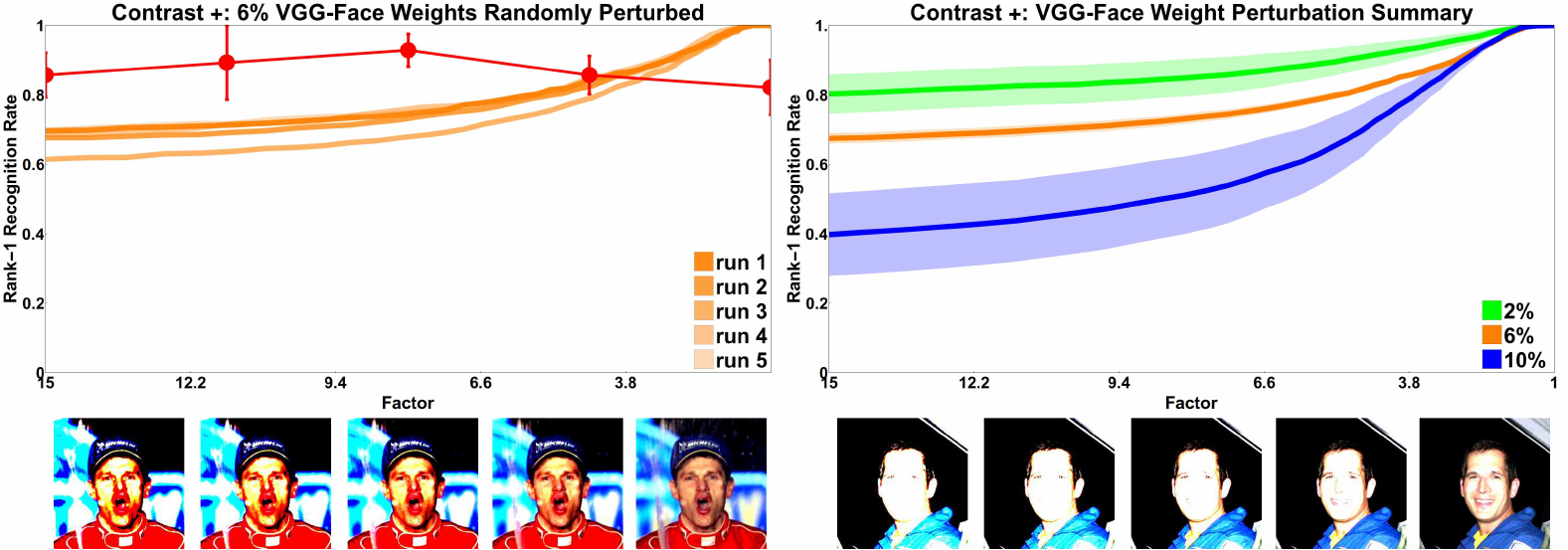}
 \caption{Weight perturbations for stochastic model output can be combined with stimulus perturbations for a stronger reliability assessment. (Left) Five independent model runs where 6\% of the weights have been perturbed, with input stimuli reflecting increasing contrast. (Right) Curves represent the average of five runs for three different levels of weight perturbation from 2\% to 10\%. Shaded regions are standard error.}
 \label{fig:weights}
\end{figure*}

\textbf{Identification with 2D Images.} Given recent results on datasets, one would expect that the deep CNNs (FaceNet, OpenFace, and VGG-Face) would be the best performers on an $M$-AFC task, following by the shallower network (slmsimple), and then the approach that makes use of handcrafted features (OpenBR). Surprisingly, this is not what we observed for any of the experiments (Figs.~\ref{fig:lfw} and~\ref{fig:composite}; Supp. Figs. 1-2). Overall, VGG-Face is the best performing network, as it is able to withstand the perturbations to a greater degree than the rest of the algorithms. At some points (\textit{e.g.}, left-hand side of Fig.~\ref{fig:lfw}) the perturbations have absolutely no effect on VGG-Face, while severely degrading the performance of other algorithms, signifying strong learned invariance. 

Remarkably, the non-deep learning approach OpenBR is not the worst performing algorithm. It turned out to outperform several of the deep networks in most experiments. This is the kind of finding that would not be apparent from a CMC or ROC curve calculated from a dataset, where OpenBR is easily outperformed by many algorithms across many datasets~\cite{openface0.2,6712754}. Why does this occur? These results indicate that it's not always possible to rely on large amounts of training data to learn strongly invariant features --- a task that can be different from learning representations that perform well on a chosen dataset. The design of the algorithm is also consequential: OpenBR's choice of LBP~\cite{1717463} and SIFT~\cite{790410} leads to better performance than FaceNet and OpenFace, which each learned features over hundreds of thousands of faces images.   

\textbf{Identification with 3D Images.} Computer graphics allows us to generate images for which all parameters are known --- something not achievable with 2D data. One such parameter, expression, has been widely studied~\cite{sim2002cmu,gross2010multi,6996248}, but not in the highly incremental manner we propose here. Where exactly do algorithms break for specific expression changes?  We can find this out by controlling the face with  graphics (Figs.~\ref{fig:fg} and~\ref{fig:composite}; Supp. Figs. 3-4). For instance, for the bodily function of blinking (Fig.~\ref{fig:fg}) VGG-Face and slmsimple are the best, while this very small change to the visual appearance of the face causes a significant degradation of matching performance in the three other algorithms. OpenFace and FaceNet once again have trouble learning invariance from their training data. This trend holds over several expressions and emotions (Supp. Figs. 3-4).

\textbf{OpenFace vs. FaceNet.} It is often difficult to assess the claims made by the developers of machine-learning-based algorithms. During the course of our experimentation, we discovered an interesting discrepancy between two networks, FaceNet~\cite{facenet} and OpenFace~\cite{amos2016openface}, which both reported to be implementations of Google's FaceNet algorithm~\cite{schroff2015facenet}. While it is good for end-users that deep learning has, in a sense, become ``plug-and-play," there is also some concern surrounding this. It is not always clear if a re-implementation of an algorithm matches the original specification. Psychophysics can help us find this out. Across all experiments, FaceNet demonstrates very weak invariance properties compared to OpenFace (Figs.~\ref{fig:fg}-\ref{fig:composite}; Supp. Figs. 3-4), and fails well before the other algorithms in most cases. From these results, we can conclude that use of this particular implementation of Google's FaceNet should be avoided. But why is it so different from OpenFace, and what would be causing it to fail, in spite of it reporting superior accuracy on LFW (0.992 for FaceNet vs. 0.9292 for OpenFace)? 

One can find three key differences in the code and data --- after being prompted to look there by the psychophysics experiments. (1) OpenFace uses 500k training images by combining CASIA-WebFace~\cite{yi2014learning} and FaceScrub~\cite{ng2014data}; FaceNet uses a subset of  MS-Celeb-1M~\cite{guo2016msceleb} where difficult images that contain partial occlusion, silhouettes, etc. \textit{have been removed} as a function of facial landmark detection. This is likely the weakest link, as the network does not have an opportunity to learn invariance to these conditions. (2) OpenFace uses the exact architecture described by Schroff et al.~\cite{schroff2015facenet}, while FaceNet opts for Inception ResNet v1~\cite{szegedy2017inception}. (3) FaceNet uses a Multi-Task CNN~\cite{zhang2016joint} for facial landmark detection and alignment, while OpenFace uses dlib~\cite{king2009dlib} --- which FaceNet intentionally avoids due to its lower yield of faces for the training set. FaceNet may have hit upon the right combination of network elements for LFW, but it does not generalize like the original work, which OpenFace is more faithful to.   

\textbf{Weight Perturbation Coupled with Stimulus Perturbation.} The procedure of applying perturbations directly to the weights of a neural network has an interpretation of Bayesian inference over the weights, and leads to stochastic output~\cite{NIPS2011_4329,goodfellow2016deep}. This is potentially important for face recognition because it gives us another measure of model reliability. To look at the effect of CNN weight perturbations coupled with stimulus perturbations, we use VGG-Face as a case study. A percentage of its weights are replaced with a random value from the normal distribution, ${\cal N}(0,1)$, targeting all layers. From Fig.~\ref{fig:weights}, we can see that both perturbation types have an impact. Under a regime that perturbs just 6\% of the weights (left-hand side of Fig.~\ref{fig:weights}), we can gain a sense that VGG-Face is stable across models with respect to its performance when processing increasing levels of contrast. However, too much weight perturbation increases the variance, leading to undesirable behavior on the perturbed input. On the right-hand side of Fig.~\ref{fig:weights}, each curve represents the average of five runs when perturbing between 2\% and 10\% of the weights. Perturbing 10\% of the weights breaks the invariant features of VGG-Face and induces more variance between models. Similar effects for other  transformations can be seen in Supp. Figs. 5-6.  

\textbf{Human Comparisons.} As discussed in Sec.~\ref{sec:related}, there is a rich literature within biometrics comparing human and algorithm performance. However, thus far, such studies have not made use of any procedures from visual psychophysics. Here we fill this gap. To obtain human data points for Figs.~\ref{fig:lfw}-\ref{fig:weights} (the red dots in the plots), we conducted
a study with $14$ participants. The task the participants performed largely followed the standard $M$-AFC protocol described above: a participant is briefly shown an image, it is hidden from sight, and then they are shown three images and directed to choose the image that is most similar to the first one. Each participant performed the task three times for each perturbation level. Each set of images within a task was chosen carefully to keep human performance from being perfect. For both 2D and 3D images, the images were divided by gender such that participants could not match solely by it~\cite{OToole1998}. For 3D images, the data was also divided by ethnicity such that it could not be the sole criterion to match by~\cite{webster2011visual}. To interrupt iconic memory~\cite{Dick1974}, after each sample image is shown, a scrambled inverse frequency function was applied to the image to produce colored noise, and shown for $500$ms prior to the alternate choices. 2D images were shown for $50$ms and 3D images for $200$ms. Consistent with previous findings~\cite{o2007face,o2008humans,o2012comparing,phillips2014comparison,phillips2015human}, we observed human performance exceeding or lagging behind algorithm performance, depending on the circumstances. Humans struggled to identify faces in the 3D context where different identities are closer in visual appearance, but excelled in the 2D context where there was greater separation between identities. The plots for Gaussian blur (Fig.~\ref{fig:lfw}) and Decreasing Contrast (Fig.~\ref{fig:composite}) hint at behavioral consistency between AI and humans in these cases.

\section{Conclusion}
 Given the model capacity of today's deep neural network-based algorithms, there is an enormous burden to explain what is learned and how that translates into algorithm behavior. Psychophysics allows us to do this in a straightforward manner when methods from psychology are adapted to conform to the typical procedures of biometric matching, as we have shown. Companies launching new products incorporating face recognition can potentially prevent (or at least mitigate) embarrassing incidents like Apple's botched demo of FaceID by matching the operational setting of an algorithm to a useable input space. And even if a company provides an explanation for a product's failure, anyone can directly interrogate it via a psychophysics experiment to find out if those claims are true. To facilitate this, all source code and data associated with this paper will be released upon publication. With the recent uptick in psychophysics work for computer vision~\cite{gerhard2013sensitive,Scheirer_2014_TPAMIa,eberhardt2016deep,richardwebster2016psyphy,DBLP:journals/corr/GeirhosJSRBW17,McCurie_2017_FG,rajalingham2018large}, we expect to see new face recognition algorithms start to use these data to improve their performance.

\clearpage

\bibliographystyle{splncs}
\bibliography{egbib}
\end{document}